\documentclass[conference]{IEEEtran}
\IEEEoverridecommandlockouts
\usepackage{cite}
\usepackage{amsmath,amssymb,amsfonts}
\usepackage{algorithmic}
\usepackage{graphicx}
\usepackage{textcomp}
\usepackage{xcolor}

\usepackage{multirow}
\usepackage{makecell}
\usepackage{rotating}
\usepackage{adjustbox}
\usepackage{caption} 

\usepackage{booktabs} 
\usepackage{multirow} 
\usepackage{graphicx} 

\usepackage{enumitem}

\def\BibTeX{{\rm B\kern-.05em{\sc i\kern-.025em b}\kern-.08em
    T\kern-.1667em\lower.7ex\hbox{E}\kern-.125emX}}
\begin{document}

\title{Unveiling Statistical Significance of Online Regression over Multiple Datasets
}

\author{\IEEEauthorblockN{1\textsuperscript{st} Mohammad Abu-Shaira}
\IEEEauthorblockA{\textit{Computer Science and Engineering} \\
\textit{The University of North Texas}\\
Denton, United States \\
shairaabu-shaira@my.unt.edu}
\and
\IEEEauthorblockN{2\textsuperscript{nd} Weishi Shi}
\IEEEauthorblockA{\textit{Computer Science and Engineering} \\
\textit{The University of North Texas}\\
Denton, United States \\
weishi.shi@unt.edu}
}

\maketitle

\begin{abstract}
Despite extensive focus on techniques for evaluating the performance of two learning algorithms on a single dataset, the critical challenge of developing statistical tests to compare multiple algorithms across various datasets has been largely overlooked in most machine learning research. Additionally, in the realm of Online Learning, ensuring statistical significance is essential to validate continuous learning processes, particularly for achieving rapid convergence and effectively managing concept drifts in a timely manner. Robust statistical methods are needed to assess the significance of performance differences as data evolves over time. This article examines the state-of-the-art online regression models and empirically evaluates several suitable tests. To compare multiple online regression models across various datasets, we employed the Friedman test along with corresponding post-hoc tests. For thorough evaluations, utilizing both real and synthetic datasets with 5-fold cross-validation and seed averaging ensures comprehensive assessment across various data subsets. Our tests generally confirmed the performance of competitive baselines as consistent with their individual reports. However, some statistical test results also indicate that there is still room for improvement in certain aspects of state-of-the-art methods. 


\end{abstract}


\section{Introduction}


In modern environments, the rapid growth of data necessitates advanced analytical tools for efficient processing and interpretation \cite{warren2015big}. Online models meet this need by incrementally updating predictive models with new data. This is especially useful in dynamic settings like finance, e-commerce, and real-time monitoring systems.

In the realm of predictive modeling, the determination of statistical significance holds paramount importance as it serves as a crucial metric for assessing the reliability and robustness of model parameters. Statistical significance not only validates the significance of predictor variables in explaining the variability of the response variable but also aids in discerning spurious relationships from genuine ones \cite{pereira2011statistically}. In dynamic environments with continuous data streams, accurately estimating statistical significance is crucial for validating model predictions amid evolving data patterns. Therefore, understanding statistical significance in online linear regression is essential for informed data-driven decision-making and improving predictive modeling.

When utilizing Machine Learning to solve problems, the selection of an appropriate model is critical. Traditional analysis relies heavily on the assumption of static datasets, where standard methods for determining statistical significance are well-established. However, these methods may not be directly applicable or adequate for online learning models due to their incremental nature and the potential non-stationarity of incoming data. 

Evaluating models is relatively straightforward when there are significant performance differences. For example, comparing two batch regression models with average \( R^2 \) values of 0.93 and 0.74 across datasets clearly identifies the superior model. However, in Online Learning, models may converge over time or adapt to concept drift, making the speed of adaptation a key focus of this research as an indicator of effectiveness. When performance differences are minimal, assessing the true superiority of one model over another requires a comprehensive analysis, considering the models' adaptability and other subtle performance metrics.


This paper aims to bridge the gap by conducting a comprehensive study on the statistical significance within online linear regression models across multiple datasets. By addressing this crucial aspect, we seek to enhance the understanding and application of online models in real-time data analysis. This investigation will provide valuable insights into the reliability and robustness of online regression techniques, contributing to their broader adoption and efficacy in handling large-scale, continuously evolving data streams. The source code and datasets utilized in this study can be accessed through our GitHub repository \cite{statistical_significance} 
and datasets repository \cite{shaira2023_olr_wa_synth_datasets}.

\section{Related Work}
Linear regression \cite{abu-mustafa2012} \cite{maulud2020review} is one of the most common and comprehensive statistical and machine learning algorithms. It establishes relationships between independent and dependent variables, making it useful for predictive analysis and identifying causal relationships in certain scenarios.

Regression may either be simple or multiple regression.
Simple linear regression studies the relationship between two continuous (quantitative) variables. One variable, denoted `x', is regarded as the predictor, explanatory, or independent variable, and the other variable, denoted `y', is regarded as the response, outcome, or dependent variable \cite{maulud2020review}. The model equation is represented by
$y= \beta_{0}  +  \beta_{1}x  + \epsilon$.
Multivariate linear regression (MLR) is used to predict a response using a number of explanatory variables. The basic model for MLR is 
$y= \beta_{0} + \beta_{1}x_{1}  + \cdots + \beta_{m}x_{m}  +   \epsilon$, where $\beta_i$'s (for i = 0, 1, ..., m) represent the coefficients associated with the respective predictor variables. Within this model, $y$ signifies the observed value of the dependent variable, while the expression $\beta_0 + \beta_{1}x_{1} + \cdots + \beta_{m}x_{m}$ denotes the anticipated or predicted value, $\epsilon$ captures the residual or error component. 


Statistical tests can be broadly categorized based on various criteria such as the type of data, the number of samples, and the nature of the hypotheses being tested \cite{lehmann1986testing}. ``All models are limited by the validity of the assumptions on which they ride'' \cite{freedman2010statistical}. Statistical techniques rely on certain underlying assumptions, with varying degrees of strictness. Sometimes, even if these assumptions are not fully met, the main findings of the research may remain unaffected. However, in other situations, failing to adhere to these assumptions can seriously weaken the validity of the research results. 

Statistical tests fall into two main categories: \textit{Parametric} and \textit{Non-parametric}. Parametric tests, which include t-tests and ANOVA, assume several conditions. First is the assumption of interval or ratio data, which is an absolute necessity for the dependent variable score; parametric statistics cannot be conducted with nominal (categorical) dependent variable data. Second is the assumption of independent scores, meaning each individual in the sample must be independent from all others. Third is the assumption of normality, where the distribution of the sample means should approximate a normal distribution, applying to the means derived from individual scores within the sample, not the individual scores themselves. Finally, the assumption of homogeneity of variance, which states that the variances of the groups being compared are equal (homoscedasticity) \cite{garson2012testing}.

In contrast, non-parametric tests do not require stringent assumptions about the population. They can be used with nominal and ordinal data and are applicable to highly skewed distributions. This makes non-parametric tests versatile for a wider range of data types, particularly when the data do not meet the strict criteria required by parametric tests. Thus, while parametric tests are powerful when their assumptions are met, non-parametric tests offer flexibility and robustness in handling various data conditions \cite{garson2012testing}.

There is a set of parametric and non-parametric tests that are suitable for assessing multi models over multi datasets. These tests are vital for rigorously comparing machine learning models across various scenarios, ensuring reliability amidst the complexity of real-world data.

\subsection{ANOVA}
The standard statistical technique for evaluating differences among more than two related sample means is the repeated-measures ANOVA, also known as within-subjects ANOVA \cite{fisher1956statistical}. ANOVA sometimes called the (Analysis of Variance) is a parametric test that helps in determining whether differences in group means are due to actual differences between groups or just random variation. Using two-way ANOVA for statistical significance in a regression context, particularly when comparing multiple online linear regression models over multiple datasets, might not be the best fit for several reasons. Firstly, ANOVA assumes equal variances (homoscedasticity) among groups. However, with datasets having varying levels of noise or variability, this assumption may be violated, leading to inaccurate results. Secondly, ANOVA assumes equal means among groups, with the null hypothesis suggesting no differences. If the p-value is below a significance threshold (often 0.05), we reject the null, implying at least one mean differs, which is not optimal for comparing multiple different models. Thirdly, ANOVA's sensitivity to outliers, often present in real-world data, can distort mean values and lead to misleading conclusions.

\subsection{Wilcoxon Signed-Rank Test}
The Wilcoxon Signed-Rank Test \cite{wilcoxon1992individual} is a non-parametric statistical test used to compare paired samples or repeated measurements from the same group, making it ideal for within-subjects designs such as pretest/posttest studies. This test operates with continuous data and is particularly valuable when the assumptions of parametric tests, like the normality of data, are not met. It is designed to assess the differences between two related groups, requiring that the paired samples are random and independent and that the distribution of the differences between groups is symmetrical. The Wilcoxon Signed-Rank Test is commonly used in scenarios where two related samples need to be compared, such as evaluating the performance of two models across multiple datasets. However, it does not extend naturally to comparisons involving more than two models \cite{demvsar2006statistical}.

\subsection{Friedman Test}
The Friedman test \cite{friedman1937use}\cite{ friedman1940comparison} is a non-parametric test that does not assume normality or equal variances, making it versatile for various types of dependent variables, including continuous ones. It is particularly useful when the assumptions of parametric tests like the repeated measures ANOVA are violated. The test ranks the performance of each model on each dataset and then compares these ranks, making it suitable for comparing multiple models on several datasets. One of its key advantages is its robustness to violations of normality and homogeneity of variances assumptions. However, it does not provide specific information about where differences lie between groups; post-hoc tests may be required for this purpose \cite{wahono2014comparison, vevcek2017influence}.

\subsection{Nemenyi's Post-hoc Test}
If the Friedman test indicates significant differences, several post-hoc tests can be applied \cite{pereira2015overview}. The Nemenyi test \cite{nemenyi1963distribution} can be used to determine which specific models differ from each other. This test helps in identifying pairs of models that have significant performance differences. The performance of two treatments is considered significantly different if the corresponding average ranks differ by at least the critical difference \cite{demvsar2006statistical}. The crucial difference of Nemenyi test is illustrated in Table \ref{tab:nemenyi_critical_difference}.

Our study encompasses a statistical significance test over the state-of-the-art online regression models. These models offer various approaches to online regression with unique strengths and considerations. Online regression via Stochastic Gradient Descent (SGD)\cite{ding2021efficient} processes one data point at a time, ideal for large datasets. Yet, it demands access to all prior data points and might converge at a slower pace. Mini-Batch Gradient Descent (MBGD)\cite{haji2021comparison} combines the strengths of batch gradient descent and SGD by updating the model with mini-batches. This approach delivers faster convergence than SGD while remaining effective for large datasets. Extending online stochastic method with L1/L2 regularization for online lasso (OLR) and ridge (ORR) regression \cite{mohri2018foundations}, reducing overfitting by penalizing model complexity. The Widrow-Hoff (LMS)\cite{widrow1960adaptive} algorithm is simple and efficient but lacks memory of past data, restricting its capability to capture long-term dependencies and intricate patterns. Recursive Least Squares (RLS)\cite{fontenla2013online} algorithm, due to its infinite memory, captures long-term dependencies more effectively and converges faster than LMS, but it demands more computational effort per iteration. Passive Aggressive (PA)\cite{crammer2006online} algorithm, closely related to SVM methods, converges quickly and operates efficiently with low time complexity. Its performance is highly dependent on the aggressiveness parameter. Lastly, Online Regression with Weighted Average (OLR-WA) is based on the weighted average of a base model which represents the already observed data and an incremental model which represents the new data. Each algorithm has its trade-offs in terms of computational efficiency, memory requirements, and adaptability to online regression learning scenarios.

\section{Method}
In linear regression, which follows the equation \( y = \beta_0 + \beta_1 x_1 + \beta_2 x_2 + \ldots + \beta_n x_n + \epsilon \), the residuals (errors) are usually assumed to be normally distributed. This assumption is expressed as \(\epsilon \sim \mathcal{N}(0, \sigma)\), indicating that the error term has a normal distribution with a mean of zero and some variance \(\sigma\). If we plot the residuals against the predictors and observe randomness with no clear pattern, it indicates that the residuals are not systematically biased, supporting the normality assumption. This is important for valid hypothesis testing and for constructing confidence intervals and prediction intervals. 

The mean of the residuals (\(\bar{\epsilon}\)) is zero or approximately zero suggests that the model is performing well, and the estimated parameters \(\beta_1, \beta_2, \ldots, \beta_n\) (slopes) and \(\beta_0\) (intercept) are reliable. The error can be expressed as \(\epsilon = y - \hat{y}\), where \(\hat{y} = \beta_0 + \beta_1 x_1 + \cdots + \beta_n x_n\). When these assumptions are met, it indicates that the linear regression model is an appropriate fit for the data.

In statistical hypothesis testing, it is crucial to precisely define the alternative hypothesis being examined. Rejecting the null hypothesis in favor of an alternative hypothesis is the common practice, but it's important to note that acceptance of a specific alternative hypothesis is not warranted due to the existence of multiple alternatives. When comparing two groups of data, the alternative hypothesis is straightforward, as it represents the contrary of the null hypothesis. For instance, when considering only two models, the null hypothesis (H$_{0}$) posits that there is no distinction between Model A and Model B. Consequently, the alternative hypothesis (H$_{1}$) suggests that Model A and Model B differ, indicating a discernible distinction. However, when dealing with multiple models, defining the alternative hypothesis becomes more complex. In this scenario, H$_{1}$ contends that there exists no disparity among Model A, Model B, Model C, etc. Nonetheless, this assertion carries ambiguity, as it could imply various interpretations, such as the differentiation of all models or distinctions solely between certain models while others remain unaffected. Additionally, P-values, ranging from 0 to 1, quantify the confidence in the differentiation between Model A and Model B. A low p-value, typically below 0.05, signifies rejection of the null hypothesis, indicating a false positive. However, while a low p-value aids in determining the dissimilarity between models, it does not provide insight into the magnitude of this difference. Thus, a small p-value does not necessarily indicate a large effect size or substantial disparity between Model A and Model B.

``Assumptions behind models are rarely articulated, let alone defended. The problem is exacerbated because journals tend to favor a mild degree of novelty in statistical procedures. Modeling, the search for significance, the preference for novelty, and the lack of interest in assumptions - these norms are likely to generate a flood of nonreproducible results'' \cite{freedman2008oasis}. In our study, our data conforms to the normality assumption required for parametric tests. However, it does not meet the assumption of Homogeneity of Variance, where the variances among the compared groups are equal (homoscedasticity). Given the presence of multiple datasets exhibiting diverse noise levels and variability, the presumption of equal variances may not hold. Consequently, parametric tests may lack the precision to appropriately account for these discrepancies.

The Friedman test, being non-parametric and capable of handling multiple treatments, three or more groups, over multiple datasets is employed to compute the test statistic and determine whether to reject the null hypothesis, thereby indicating differences among the models. Subsequently, the Nemenyi post-hoc test is utilized to verify if each model significantly differs in performance from its counterparts. We utilize both  Critical Difference values, CD$_{\alpha=0.05}$ and CD$_{\alpha=0.1}$, to enable a comprehensive evaluation of statistical significance across varying levels of stringency. This approach facilitates a nuanced interpretation of the results, accommodating both highly significant findings and those of marginal significance results.


In the domain of online learning, data arrives incrementally, leading models to gradually converge towards discernible data patterns over successive iterations, typically referred to as mini-batches. Through measuring the Mean Square Error (MSE) across all iterations, our analysis reveals that all models ultimately converge, exhibiting nearly equivalent performance levels in later stages of data intake. However, disparities among models surface initially, showcasing discrepancies in their rates of convergence, which in turn reflects their respective capacities to adapt to concept drift, a common occurrence in online learning. Table \ref{tab:datasets-properties} showcases the application of eight distinct real and synthetic datasets, encompassing a wide array of scenarios, including low and high dimensionality, varying sample sizes, and levels of noise. Furthermore, comprehensive evaluation is conducted through the utilization of 5-fold cross-validation combined with an average seeding technique.

\section{Experiments}

Our experiments entail the comparison of eight online linear regression models.Table \ref{table:models_performance_first_10_minibatches} illustrates the performance of the models computed on the initial ten mini-batches, as measured by Mean Squared Error (MSE), and presented as an averaged score. The batch size varies according to the characteristics of each dataset. The performance results serve as the input for the conducted statistical tests.

The non-parametric Friedman test is utilized to compare the multiple models. In our example, we want to compare the performance of k=8 different models, on N=8 datasets. First, the test calculates the average rank of each model’s performance on each dataset, with the best-performing model receiving a rank of 1. The Friedman test then tests the null hypothesis,$ H_0$, that all models are equally effective and their average ranks should be equal. The Friedman test statistic is calculated using Equation \ref{eq:friedman_equation}, where $R$ is the average ranking of each model.

\begin{equation}
\chi_{F}^{2}=\frac{12 N}{k(k+1)}\left[\sum_{j=1}^{k} R_{j}^{2}-\frac{k(k+1)^{2}}{4}\right]   
\label{eq:friedman_equation}
\end{equation}

The test result can be used to determine whether there is a statistically significant difference between the performance of the models by making sure that $\chi_{F}^{2}$ is not less than the critical value for the F distribution for a particular confidence value $\alpha$. However, since $\chi_{F}^{2}$ could be too conservative, we also calculate the $F_F$ statistic using Equation \ref{eq:ff_statistic_equation}.

\begin{equation}
F_{F}=\frac{(N-1) \chi_{F}^{2}}{N(k-1)-\chi_{F}^{2}}
\label{eq:ff_statistic_equation}
\end{equation}

\begin{align*}
\chi_{F}^{2} = & \frac{12 \times 8}{8 \times 9} \left[ (5.125^2 + 7.625^2 + 4.375^2 + 5^2 + 3.875^2 \right. \\
& \left. + 4.875^2  + 3.875^2 + 1.250^2) - \frac{8 \times 9^{2}}{4} \right] = 29.20
\end{align*}

\[
F_{F}=\frac{(7) \times 29.20}{8 \times 7 - 29.20} = 7.63
\]

The critical value can be determined from the F-distribution table. In this table, the degrees of freedom across columns (represented as \( df_1 \)) correspond to \( k-1 \), where \( k \) denotes the number of models minus one. Similarly, the degrees of freedom across rows (denoted as \( df_2 \)) equate to \( (k-1) \times (N-1) \), indicating the product of the number of models minus one and the number of datasets minus one. In our specific scenario, \( df_1 = 7 \) and \( df_2 = 49 \). Since \( df_2 = 49 \) does not directly appear in the F Table, we resort to linear interpolation. The interpolation formula is expressed as:

\[ F_{df_2 = 49} = F_{40} + \left(\frac{49 - 40}{60 - 40}\right) \times (F_{60} - F_{40}) \]



\[ F_{df_2 = 49} = 2.2490 + \left(\frac{49 - 40}{60 - 40}\right) \times (2.1665 - 2.2490) \]

\[ = 2.21187\]

With eight algorithms and eight data sets, $F_{F}$ is distributed according to the F distribution with
$8 - 1 = 7$ and $(8 - 1) \times (8 - 1) = 49$ degrees of freedom. The critical value of F(7,49) for $\alpha$ = 0.05
is 3.04705. Since the critical value is below our obtained statistic, so we reject the null-hypothesis with 95\% confidence.

Following the rejection of the null hypothesis, a post hoc test is performed. Specifically, the Nemenyi test is employed to ascertain significant differences in the performance of models. The null hypothesis \( H_0 \) for the Nemenyi  test posits that there is no difference between any two models, and the alternative hypothesis \( H_1 \), indicating that at least one pair of models displays a discernible difference. The Critical Difference (CD) is defined by Equation \ref{eq:nemenyi_cd}.

\begin{equation}
CD = q_{\alpha} \sqrt{\frac{k(k+1)}{6N}}
\label{eq:nemenyi_cd}
\end{equation}

where $q_{\alpha}$ is the critical difference of the studentized range distribution at the chosen significance level and k is the number of groups. The $q_{\alpha}$ value can be obtained using Table \ref{tab:nemenyi_critical_difference}. Therefore, for our specific case study, the critical differences are:

\[CD_{\alpha=.05} = 3.031 \sqrt{\frac{8 \times 9}{6 \times 8}} = 3.712\] 

\[CD_{\alpha=.1} = 2.780 \sqrt{\frac{8 \times 9}{6 \times 8}} = 3.404\]

\begin{table}[t]
\centering
\captionsetup{justification=centering}
\caption{Dataset Properties.\\ {\scriptsize Incorporated 5-folds cross-validation with seed averaging}}
\renewcommand{\theadalign}{bc} 
\renewcommand{\theadfont}{\bfseries\scriptsize} 
\setlength{\tabcolsep}{3.5pt} 
\begin{tabular}{cccccccc}
\toprule
\rotatebox{90}{\thead{Dataset}} & 
\rotatebox{90}{\thead{Type}} & 
\rotatebox{90}{\thead{Data-\\points}} & 
\rotatebox{90}{\thead{Dimen-\\sions}} & 
\rotatebox{90}{\thead{Noise}} & 
\rotatebox{90}{\thead{Train\\(5-Fold)}} & 
\rotatebox{90}{\thead{Test\\(5-Fold)}} &
\rotatebox{90}{\thead{Batch\\Size}} \\
\midrule
DS1 \cite{shaira2023_olr_wa_synth_datasets} & Synth. & 1k & 3 & 10 & 800 & 200 & 10  \\
DS2 \cite{shaira2023_olr_wa_synth_datasets} & Synth. & 10k & 20 & 20 & 8k & 2k & 30 \\
DS3 \cite{shaira2023_olr_wa_synth_datasets} & Synth. & 10k & 200 & 25 & 8k & 2k  & 300\\
DS4 \cite{shaira2023_olr_wa_synth_datasets} & Synth. & 50k & 500 & 50 & 40k & 10k  & 1k\\
MCPD \cite{MCPD_DS} & Real & 1.3k & 7 & N.A & 1.1k & 267  & 25\\
1KC \cite{1KC_DS} & Real & 1k & 5 & N.A & 800 & 200 & 30\\
KCHSD \cite{KCHS_DS} & Real & 21.6k & 21 & N.A & 17.3k & 4.3k & 30 \\
CCPP \cite{CCPP_DS} & Real & 9.6k & 5 & N.A & 7.7k & 1.9k  & 20\\
\bottomrule
\end{tabular}
\label{tab:datasets-properties}
\end{table}

\begin{table*}[t]
\captionsetup{justification=centering}
\caption{Performance Analysis on Online Regression Scenarios\\
{\scriptsize Summary of Performance Measures Using MSE on the First 10 Iterations; \quad  
\scriptsize E: epochs, $\mathbb{Z}^{+}$: Natural Number $\geq$ 1}}
\centering
\renewcommand{\theadalign}{cc} 
\renewcommand{\theadfont}{\small} 
\setlength{\tabcolsep}{6pt} 
\begin{adjustbox}{width=\textwidth} 
\begin{tabular}{|c|*{16}{c|}}
\hline
\multirow{3}{*}{\thead{Datasets}} & \multicolumn{16}{c|}{\thead{Algorithms}} \\ \cline{2-17} 
& \multicolumn{2}{c|}{\rotatebox[origin=c]{90}{\thead{\textbf{SGD} \\ {\scriptsize $\eta=0.01$ [DS3, DS4 =0.001},\\ \scriptsize{ KCHSD =$1 \mathrm{e}{-4}$]}, \\ {\scriptsize E=N $\times$ $(\mathbb{Z}^{+}=2)$ }\\{\scriptsize [CCPP $\mathbb{Z}^{+}=5$]}}}} 
& \multicolumn{2}{c|}{\rotatebox[origin=c]{90}{\thead{\textbf{MBGD} \\ {\scriptsize $\eta=0.01$ [1KC= 0.1],}\\{\scriptsize$ E=\frac{N}{K} \times (\mathbb{Z}^{+} =5)$ [DS3, DS4 \vspace{1.5mm}}\\{\scriptsize  $\mathbb{Z}^{+}$=100, KCHSD $\mathbb{Z}^{+}$=20,}\\{\scriptsize 1KC, CCPP $\mathbb{Z}^{+}$=40]}}}} 
& \multicolumn{2}{c|}{\rotatebox[origin=c]{90}{\thead{\textbf{LMS} \\ {\scriptsize $\eta=0.01$[DS3, DS4=0.001},\\ \scriptsize{1KC=.1, KCHSD,CCPP=$1 \mathrm{e}{-4}$]}}}} 
& \multicolumn{2}{c|}{\rotatebox[origin=c]{90}{\thead{\textbf{ORR} \\ {\scriptsize $\eta=0.01$ [DS3, DS4, KCHSD,}\\ \scriptsize{=0.001], E=N $\times (\mathbb{Z}^{+}=2)$}\\{\scriptsize [1KC $\mathbb{Z}^{+}$=3, KCHSD, CCPP}\\ \scriptsize{ $\mathbb{Z}^{+}=5$], $\lambda=0.1$ [1KC=0.01,}\\{\scriptsize MCPD,KCHSD,CCPP=0.001]}}}} 
& \multicolumn{2}{c|}{\rotatebox[origin=c]{90}{\thead{\textbf{OLR} \\ {\scriptsize $\eta=0.01$ [DS3, DS4, KCHSD}\\{\scriptsize = 0.001], E=N $\times (\mathbb{Z}^{+}=2)$}\\{\scriptsize [1KC $\mathbb{Z}^{+}$=3, KCHSD, CCPP }\\{\scriptsize $\lambda=0.1$ [1KC, CCPP=0.01],}\\ \scriptsize{[ KCHSD=$1 \mathrm{e}{-4}$]}}}} 
& \multicolumn{2}{c|}{\rotatebox[origin=c]{90}{\thead{\textbf{RLS} \\ {\scriptsize $\lambda=.99$, $\delta=0.01$}\\{\scriptsize [1KC $\delta=0.1$]}}}}        
& \multicolumn{2}{c|}{\rotatebox[origin=c]{90}{\thead{\textbf{PA} \\ {\scriptsize \text{PA-}\uppercase\expandafter{\romannumeral3}}\\ {\scriptsize $C=0.1$, $\epsilon=0.1$},\\ \scriptsize{ [DS6, DS7, DS8}\\ \scriptsize{$C=0.01$, $\epsilon=0.01$]}}}} 
& \multicolumn{2}{c|}{\rotatebox[origin=c]{90}{\thead{\textbf{OLR-WA} \\ {\scriptsize W$_{base}=.5$, W$_{inc}=.5$}\\{\scriptsize [KCHSD W$_{base}=.9$,}\\{\scriptsize W$_{inc}=.1$]}}}} \\ \cline{2-17} 
& \thead{MSE} & \thead{rank} & \thead{MSE} & \thead{rank} & \thead{MSE} & \thead{rank} & \thead{MSE} & \thead{rank} & \thead{MSE} & \thead{rank} & \thead{MSE} & \thead{rank} & \thead{MSE} & \thead{rank} & \thead{MSE} & \thead{rank} \\ \hline
\thead{DS1 \cite{shaira2023_olr_wa_synth_datasets}} & 1865.20581 & 4 & 5976.67276 & 8 & 1778.39859 & 3 & 1961.97400 & 6 & 1933.74953 & 5 & 3080.20632 & 7 & 541.27625 & 2 & 455.90216 & 1 \\ \hline
\thead{DS2 \cite{shaira2023_olr_wa_synth_datasets}} & 2430.97513 & 5 & 23760.09556 & 8 & 2390.53622 & 4 & 2825.94930 & 6 & 2122.09717 & 3 & 4715.74581 & 7 & 1564.91181 & 2 & 841.70463 & 1 \\ \hline
\thead{DS3 \cite{shaira2023_olr_wa_synth_datasets}} & 3282.66236 & 6 & 31822.14283 & 8 & 3275.53508 & 5 & 3887.32562 & 7 & 3218.66053 & 4 & 1593.63189 & 2 & 2349.97800 & 3 & 1165.10646 & 1 \\ \hline
\thead{DS4 \cite{shaira2023_olr_wa_synth_datasets}} & 5550.43115 & 5 & 31435.32987 & 8 & 5547.42274 & 4 & 5658.62652 & 6 & 5429.84169 & 3 & 12569.60925 & 7 & 5407.04210 & 2 & 3565.32212 & 1 \\ \hline
\thead{MCPD \cite{MCPD_DS}} & 0.33973 & 8 & 0.85781 & 8 & 0.31242 & 1 & 0.33924 & 3 & 0.33995 & 5 & 0.36409 & 7 & 0.36793 & 7 & 0.31687 & 2 \\ \hline
\thead{1KC \cite{1KC_DS}} & 0.00362 & 6 & 0.00709 & 8 & 0.00138 & 2 & 0.00358 & 4 & 0.00359 & 5 & 0.00151 & 3 & 0.00464 & 7 & 0.00097 & 1 \\ \hline
\thead{KCHSD \cite{KCHS_DS}} & 0.88983 & 7 & 0.69907 & 6 & 0.89519 & 8 & 0.59394 & 5 & 0.57859 & 4 & 0.49323 & 1 & 0.50757 & 3 & 0.50742 & 2 \\ \hline
\thead{CCPP \cite{CCPP_DS}} & 0.05089 & 4 & 0.18292 & 7 & 0.23790 & 8 & 0.05065 & 3 & 0.05058 & 2 & 0.09088 & 6 & 0.07098 & 5 & 0.00407 & 1 \\ \hline

\hline

\thead{\textbf{Average}} & 
1,641.32 & 5.125 &
11,624.50& 7.625 &
1624.16744& 4.375 &
1,791.86& 5 &
1,588.17& 3.875 &
2,745.02& 4.875 &
1,233.02& 3.875 &
753.61  & 1.250 \\ \hline
\end{tabular}
\end{adjustbox}
\label{table:models_performance_first_10_minibatches}
\end{table*}

\begin{table}[h]
\caption{Critical Values for Nemenyi Test \cite{demvsar2006statistical}}
\setlength{\tabcolsep}{5.5pt} 
\centering
\begin{tabular}{|c|c|c|c|c|c|c|c|c|}
\hline
\# Models & 2 & 3 & 4 & 5 & 6 & 7 & 8 \\
\hline
$\alpha$=.05 & 1.960 & 2.343 & 2.569 & 2.728 & 2.850 & 2.948 & 3.031 \\
\hline
$\alpha$=0.1 & 1.645 & 2.052 & 2.291 & 2.459 & 2.589 & 2.693 & 2.780 \\
\hline
\end{tabular}
\label{tab:nemenyi_critical_difference}
\end{table}

In the Nemenyi test, the Critical Difference (CD) denotes the minimum average rank difference necessary between two models to deem them significantly different. When the discrepancy in average ranks exceeds the CD, we can reject the null hypothesis, indicating a significant disparity in model performances.

\begin{table}[h]
\centering
\caption{Absolute Differences in Average Ranks}
\setlength{\tabcolsep}{5pt} 
\begin{tabular}{|c|c|c|c|c|c|c|c|}
\hline
\textbf{Model} & \textbf{SGD} & \textbf{MBGD} & \textbf{LMS} & \textbf{ORR} & \textbf{OLR} & \textbf{RLS} & \textbf{PA} \\ \hline
\textbf{SGD} & - & 2.5 & 0.75 & 0.125 & 1.25 & 0.25 & 1.25 \\ \hline
\textbf{MBGD} & 2.5 & - & 3.25 & 2.625 & 3.75 & 2.75 & 6.375 \\ \hline
\textbf{LMS} & 0.75 & 3.25 & - & 0.625 & 0.5 & 0.5 & 1.5 \\ \hline
\textbf{ORR} & 0.125 & 2.625 & 0.625 & - & 1.875 & 1.125 & 3.625 \\ \hline
\textbf{OLR} & 1.25 & 3.75 & 0.5 & 1.875 & - & 0.25 & 2.625 \\ \hline
\textbf{RLS} & 0.25 & 2.75 & 0.5 & 1.125 & 0.25 & - & 2.625 \\ \hline
\textbf{PA} & 1.25 & 6.375 & 1.5 & 3.625 & 2.625 & 2.625 & - \\ \hline
\textbf{OLR-WA} & 3.875 & 6.375 & 3.125 & 3.750 & 2.625 & 3.625 & 2.625 \\ \hline
\end{tabular}
\label{tab:nem_res}
\end{table}

Table \ref{tab:nem_res} displays the calculated absolute differences in average ranks among the models. The average rank used by Friedman's and Nemenyi's tests consequently serves as a crucial metric for assessing the overall performance of each model across multiple datasets, providing valuable insights into their relative effectiveness. In this study, we'll focus our analysis on OLR-WA due to its attainment of the lowest average rank among the competing models evaluated in our study, facilitating a comprehensive understanding of the applicability and potential benefits in real-world regression scenarios.

\begin{enumerate}[align=left, leftmargin=10pt, labelwidth=0pt]
\item OLR-WA vs. SGD:
The absolute difference in the average rank between OLR-WA and SGD is
3.875, which is higher than the CD$_{\alpha=.05} = 3.712 $ and CD$_{\alpha = 0.1} = 3.404$ respectively, thus, the observed absolute difference exceeds both CD values, signifying a statistically significant difference in performance between OLR-WA and SGD. OLR-WA outperforms SGD with a lower average rank, indicating superior performance across the datasets.

\item OLR-WA vs. MBGD:
The absolute difference in the average rank between OLR-WA and MBGD is
6.375, which is higher than the CD$_{\alpha=.05} = 3.712 $ and CD$_{\alpha = 0.1} = 3.404$ respectively, thus, the observed absolute difference exceeds both CD values, signifying a statistically significant difference in performance between OLR-WA and MBGD. OLR-WA outperforms MBGD with a lower average rank, indicating superior performance across the datasets.

\item OLR-WA vs. LMS: 
The absolute difference in the average rank between OLR-WA and LMS is
3.125, which is not higher than the CD$_{\alpha=.05} = 3.712 $ and CD$_{\alpha = 0.1} = 3.404$ respectively, thus, the absolute difference in average ranks does not exceed either CD value, indicating that the observed difference in performance between OLR-WA and LMS is not statistically significant. Consequently, there is insufficient evidence to conclude that OLR-WA performs differently from LMS based on the Nemenyi test.

\item OLR-WA vs. ORR:
The absolute difference in the average rank between OLR-WA and ORR is
3.750, which is higher than the CD$_{\alpha=.05} = 3.712 $ and CD$_{\alpha = 0.1} = 3.404$ respectively, thus, the observed absolute difference exceeds both CD values, signifying a statistically significant difference in performance between OLR-WA and ORR. OLR-WA outperforms ORR with a lower average rank, indicating superior performance across the datasets.


\item OLR-WA vs. OLR: 
The absolute difference in the average rank between OLR-WA and OLR is
2.625, which is not higher than the CD$_{\alpha=.05} = 3.712 $ and CD$_{\alpha = 0.1} = 3.404$ respectively, thus, the absolute difference in average ranks does not exceed either CD value, indicating that the observed difference in performance between OLR-WA and OLR is not statistically significant. Consequently, there is insufficient evidence to conclude that OLR-WA performs differently from LMS based on the Nemenyi test.

\item OLR-WA vs. RLS: 
The absolute difference in the average rank between OLR-WA and RLS is
3.625, which is not higher than the CD$_{\alpha=.05} = 3.712 $ and CD$_{\alpha = 0.1} = 3.404$ respectively, thus, the absolute difference in average ranks does not exceed either CD value, indicating that the observed difference in performance between OLR-WA and RLS is not statistically significant. Consequently, there is insufficient evidence to conclude that OLR-WA performs differently from RLS based on the Nemenyi test.

\item OLR-WA vs. PA: 
The absolute difference in the average rank between OLR-WA and PA is
2.625, which is not higher than the CD$_{\alpha=.05} = 3.712 $ and CD$_{\alpha = 0.1} = 3.404$ respectively, thus, the absolute difference in average ranks does not exceed either CD value, indicating that the observed difference in performance between OLR-WA and PA is not statistically significant. Consequently, there is insufficient evidence to conclude that OLR-WA performs differently from PA based on the Nemenyi test.

\end{enumerate}

\section{Conclusion}
This study examines the statistical significance of state-of-the-art online regression models across a range of real and synthetic datasets with employing 5-folds cross validation and seed averaging for comprehensive evaluation across different subsets of the data. Employing the Friedman's test alongside its corresponding Nemenyi's post-hoc analysis, our findings reveal disparities in model performance, leading to the rejection of the null hypothesis. Specifically, the OLR-WA model demonstrates the lowest average rank of 1.25. However, upon conducting the Nemenyi test and considering the absolute difference in average ranks alongside critical difference values, it becomes apparent that OLR-WA outperforms SGD, MBGD, and ORR. Moreover, despite its superior average rank, the observed performance difference between OLR-WA and LMS, OLR, and PA does not surpass the CD values, indicating a lack of statistical significance. Consequently, there is insufficient evidence to conclude that OLR-WA performs differently from each of LMS, OLR, and PA based on the Nemenyi test.

Researchers should care about these findings as they underscore the importance of selecting models based on data-specific characteristics rather than a one-size-fits-all approach. Understanding the factors influencing model performance could lead to the development of adaptive models capable of handling rapid convergence and concept drift, performing well across diverse datasets. This nuanced understanding aids in better decision-making for more accurate and reliable predictive models in real-world applications.


\bibliographystyle{IEEEtran}
\bibliography{references}
\end{document}